\documentclass[letterpaper, 10 pt, conference]{ieeeconf}  

\IEEEoverridecommandlockouts                              

\usepackage{cite}
\usepackage{amsmath,amssymb,amsfonts}
\usepackage{algorithmic}
\usepackage{graphicx}
\usepackage{textcomp}
\usepackage{subcaption}
\usepackage{url}
\usepackage{xcolor}
\usepackage{multirow}
\usepackage{balance}

\title{\LARGE \bf
Towards Multimodal Social Conversations with Robots:\\Using Vision-Language Models
}

\author{Ruben Janssens and Tony Belpaeme
\thanks{This research received funding from the Flemish Government (AI Research Program 2).}
\thanks{Authors are affiliated with IDLab-AIRO, Ghent University--imec, Ghent, Belgium. Contact: 
        {\tt\small ruben.janssens@ugent.be}}%
}

\begin{document}

\maketitle
\thispagestyle{empty}
\pagestyle{empty}

\begin{abstract}
Large language models have given social robots the ability to autonomously engage in open-domain conversations. However, they are still missing a fundamental social skill: making use of the multiple modalities that carry social interactions. While previous work has focused on task-oriented interactions that require referencing the environment or specific phenomena in social interactions such as dialogue breakdowns, we outline the overall needs of a multimodal system for social conversations with robots.
We then argue that vision-language models are able to process this wide range of visual information in a sufficiently general manner for autonomous social robots. We describe how to adapt them to this setting, which technical challenges remain, and briefly discuss evaluation practices.
\end{abstract}



\section{Introduction}



Spoken dialogue is arguably the most advanced bidirectional communication method we have as humans. It enables us to exchange information quickly, but more importantly, it is the cornerstone of \textit{social interactions}.
Social interactions serve a higher purpose than merely exchanging factual information: they enable the building of bonds between social actors, and this is largely driven by \textit{social dialogue} \cite{zegarac1998phatic}. 

It is important that social robots can engage in social dialogue. Social robots are used in socially assistive scenarios, such as being a companion, where they should be able to build a bond with their user, or language tutoring, where they should be able to prepare their learner for human-human dialogue.
However, social robots are currently not yet able to fully engage in social dialogue. They are still missing fundamental social skills.

After all, what makes spoken dialogue so powerful and efficient, is that it uses multiple modalities. Not only the words we speak convey information, but we also use non-verbal communication, transmitted through audible and visible signals. These signals convey information about the cognitive and affective state of the conversation partner, about the identity of the conversation partner, and about the environment---all of which allows us, crucially, to \textit{adapt} to the conversation partner.



While large language models (LLMs) have allowed social robots to finally engage autonomously in verbal interactions that are virtually unconstrained \cite{pinto2025designing}, they are not yet able to make use of these multiple modalities.
Vision-language models (VLMs) promise to jointly process visual input with text in dialogue. However, they are generally not designed for social conversations with an embodied agent that is situated in the physical world.
%
Instead, they are designed to be assistants, which answer questions based on additional visual context, as in an image sent through a chat message.

Previous work has already explored the use of multimodal information in human-robot conversations---it is hardly a new challenge \cite{nagao1994social}. 
However, previous work either did not yet use language models, which inherently limited its conversational abilities \cite{fernandez2020modelling, jokinen2013multimodal}; limited its representation of visual information to discrete categories e.g. emotions or specific visual cues \cite{addlesee2024multi}; or it did not focus on social conversations, but rather tackled task-oriented conversations where visual grounding referred to objects that are used in a task \cite{allgeuer2024robots}; or focused on the start of a social conversation \cite{janssens2024integrating, pantazopoulos2021vica}.


In this position paper, we propose to explore the capabilities of VLMs to engage in \textit{situated social conversations} for robots. We will outline what capabilities are required for this system and propose an architecture for such a system, highlighting technical challenges along the way. Finally, we discuss how such a system can be evaluated. 
We note that while audible signals are a very important part of non-verbal communication, we are currently not yet aware of audio description models that can represent these in a sufficiently general way, and therefore focus on visual input.






\begin{figure}
    \centering
    \includegraphics[width=0.9\columnwidth]{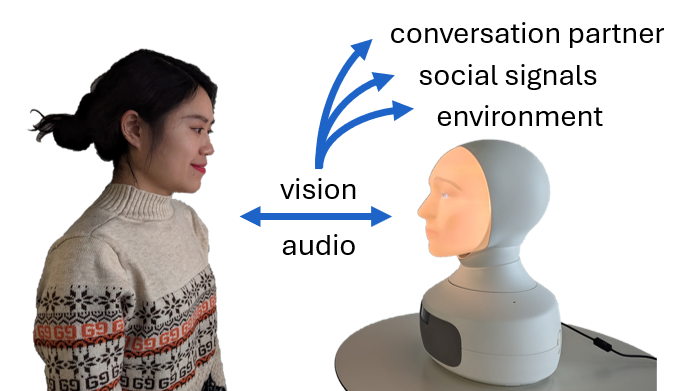}
    \caption{Overview of important information conveyed through visual and auditive channels during social interactions.\vspace{-0.5cm}}
    \label{fig:overview}
\end{figure}


\section{System Requirements and Challenges}

\begin{figure}
    \centering
    \includegraphics[width=\columnwidth]{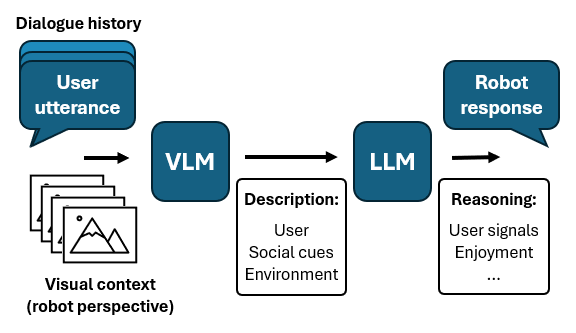}
    \caption{Proposed architecture of a multimodal adaptive social dialogue system for a social robot using a VLM and LLM.\vspace{-0.5cm}}
    \label{fig:overview-system}
\end{figure}

We outline three categories of information that a social robot should be able to process from non-verbal signals in social dialogue, shown in Figure~\ref{fig:overview}.
%
First, it should be able to detect the \textit{identity of the conversation partner}, possibly recognising them, but also adapting the conversation to them (such as how one will speak differently to a child than to an older adult), or making references to their behaviour or appearance where appropriate (e.g. recognising that the user is wearing a T-shirt with a particular band name, or that the user is drinking from a mug).
Second, it should be able to \textit{recognize and interpret social cues and signals}. 
Using these cues, the robot can build a mental model of the user and of their emotions and cognitive processes, and adapt to them.
This can help the robot to make the conversation more enjoyable for the user.
Additionally, it is an essential capability for the robot to recognise the user's feedback through these multi-modal social cues
 \cite{axelsson2022modeling}, 
 potentially repairing dialogue breakdowns. In particular in educational applications such as language learning, the robot should keep track of whether the user understands its speech, in order to model and adapt to the learner's proficiency.
Finally, it should \textit{understand the environment} the interaction is taking place in, and be able to make and understand references to that environment.


Processing this information is challenging. Social cues in particular are known to be subtle and fast \cite{krumhuber2013effects}, highly individual \cite{d2018affective}, and context-dependent \cite{mesquita2014emotions}. While existing models are able to recognise or quantify facial expressions, pose, and gestures, they are not able to interpret them further than e.g. basic categories or representations of emotions.
Furthermore, the vision model should be sufficiently general to work in any environment and with any user---not exhibiting harmful biases. Finally, visual recordings of users, their non-verbal expressions, and their environment, are personal data and should be processed and stored with respect for the user's privacy, particularly in sensitive applications, e.g. with vulnerable users such as children or older adults.

The current generation of VLMs shows promising performance on a large variety of vision-language tasks \cite{zhang2024vision} including integrating vision in conversations \cite{abbo2025blind}. Combined with their growing context windows \cite{grattafiori2024llama} (allowing the processing of many images at once), as well as high-resolution picture input \cite{gemini35}, we argue they could be able to process and represent the information outlined in Figure~\ref{fig:overview}. We propose to test whether they are able to do so, and whether they are able to interpret these signals to successfully adapt the conversation to these signals using the ``world knowledge'' they contain. While they are not trained to process visual input from an egocentric, embodied perspective, they might be able to interpret the visual information in such a scenario through a multi-step approach, leveraging the descriptive and reasoning capabilities of VLMs and LLMs. A proposal architecture for this multi-step approach is shown in Figure~\ref{fig:overview-system}.

Besides the interpretation of these cues, processing them still poses significant technical challenges. As social cues can be very short-lasting, such as microexpressions, visual information needs to be processed at a high framerate. This might necessitate intelligent sampling of these frames from the video, e.g. based on relevant movements. Furthermore, the visual information in the frames of the video needs to be aligned with the words that are spoken at that time. Finally, all processing and generation steps in the pipeline need to be executed in a sufficiently short delay---generally, one second is viewed as a maximum delay in order to not hinder a real-time spoken dialogue \cite{skantze2021turn}.

While we propose first evaluating the performance of VLMs and LLMs in this setting using zero-shot prompting or in-context learning, it could be valuable to fine-tune these models. However, the availability of data is a limiting factor, as well as the design of an appropriate loss function, as there is not one ``ground truth'' answer in a social conversation nor a suitable metric that can be automatically calculated \cite{janssens2022cool}.








\section{Evaluation}








Evaluating the performance of a social dialogue system for social robots is a difficult task and there is not yet an agreed-upon standard practice. In conversations without a particular goal besides having a social interaction, the topic of the conversation could be an important influencing factor on the ``quality'' of the conversation, and we are essentially evaluating an open-domain conversational system. Such systems can be evaluated on large-scale datasets of conversations gathered from real-world deployments of the system \cite{venkatesh2018evaluating}, which is difficult to obtain with social robots. Nevertheless, a dataset of conversations with this system could be rated, either on a turn-by-turn basis or for the whole dialogue, by the user themselves \cite{hoffman2013effects} or an external annotator \cite{raj2020defining}. This rating could be performed for various concepts, such as the user's experience of the conversation, e.g. their enjoyment, or the user's perceptions of the robot. More detailed phenomena could also be rated \cite{paetzel2024automatic}, such as the presence of dialogue loops, self-disclosure by the user, dialogue breakdowns and their repairs, the accuracy of factual information, or the correctness of visual references.
In particular regarding visual input, it is still unclear how to design a reproducible experiment that can evaluate the processing of visual cues in a social setting, going beyond references to visual objects or scripted conversation failures.
Taking inspiration from evaluation practices for LLMs, evaluation could also be done by simple preference ratings between a pair of potential responses, and ratings could be performed by a state-of-the-art LLM acting as an ``LLM-as-a-judge'' \cite{zheng2023judging, pereira2024multimodal, janssens2025online}.
Finally, objective metrics, such as the number of turns in the conversation and the length of the turns, can also provide useful information \cite{paetzel2024automatic}.
With this short discussion, we aim to prompt a discussion in the community about good practices in the unique field of multimodal social human-robot dialogue.




\balance
\bibliographystyle{IEEEtran}
\bibliography{lib}

\end{document}